\documentclass{article}
\usepackage{jfrExamplee}
\usepackage{graphicx}
\usepackage{apalike}
\usepackage{setspace}

\title{An Analysis of International Use of Robots for COVID-19}
\author{
Robin R. Murphy \\
Depart. of Computer Science \& Engineering \\
Texas A\&M University \\
College Station, TX 77843-3112 \\
\texttt{robin.r.murphy@tamu.edu}\\
\And
Vignesh B.M. Gandudi \\
Depart. of Computer Science \& Engineering \\
Texas A\&M University \\
College Station, TX 77843-3112 \\
\texttt{vigneshbabu.gm@tamu.edu}\\
\AND
Trisha Amin \\
School of Public Health \\
Texas A\&M University \\
College Station, TX 77843-3112 \\
\texttt{trishaamin@tamu.edu}\\
\And
Angela Clendenin \\
Department of Epidemiology \\
Biostatistics \\
Texas A\&M University \\
College Station, TX 77843-3112 \\
\texttt{clendenin@tamu.edu}\\
\AND
Jason Moats \\
Texas A\&M Engineering Extension Service \\
College Station, TX 77843-3112 \\
\texttt{jason.moats@teex.tamu.edu}}

\begin{document}
\maketitle
\begin{abstract}

This article analyses data collected on 338 instances of robots used explicitly in response to COVID-19 from 24 Jan, 2020, to 23 Jan, 2021, in 48 countries. 
The analysis was guided by four overarching questions: 1) What were robots used for in the COVID-19 response?  2) When were they used? 3) How did different countries innovate? and 4) Did having a national policy on robotics influence a country's innovation and insertion of robotics for COVID-19? The findings indicate that robots were used for six different sociotechnical work domains and 29 discrete use cases. When robots were used varied greatly on the country; although many countries did report an increase at the beginning of their first surge. To understand the findings of how innovation occurred, 
the data was examined through the lens of the technology's maturity according to NASA's Technical Readiness Assessment metrics. Through this lens, findings note that existing robots were used for more than 78\% of the instances; slightly modified robots made up 10\%; and truly novel robots or novel use cases constituted 12\% of the instances. The findings clearly indicate that countries with a national robotics initiative were more likely to use robotics more often and for broader purposes. Finally, the dataset and analysis produces a broad set of implications that warrant further study and investigation. The results from this analysis are expected to be of value to the robotics and robotics policy community in preparing robots for rapid insertion into future disasters.

\end{abstract}

\section{Introduction}
\label{sec:introduction}

The COVID-19 pandemic provides an unique opportunity to examine the in situ operationalization of robots
during a pandemic, and, by extension, disasters in general. Several articles have already been published that attempt to summarize how robots have been widely used to mitigate
the medical, economic, and social impacts of COVID-19, notably \cite{conversation,murphy:SSRR2020,shen2020}. More often, survey articles such as 
\cite{clipper:2020,elavarason:2020,mardani:2020,yang20}
focus on how robots could be used 
and on technological gaps or deficiencies preventing such uses.

While such analyses are valuable to the robotics community, they neglect larger questions as to whether there are patterns of how different countries have employed robots to cope with COVID-19. 
Establishing which robots have been used for what purposes and how quickly they were deployed provides a baseline for future work. 
Analyzing the
influences on why and how those robots were put into practice
is valuable in determining research investments and national policies to prepare for effective response to future disasters. 

In order to provide this larger view of the use of robotics for COVID-19, this article surveys the state of the practice and addresses four specific questions:

\begin{itemize}
    \item \textit{What were robots used for?} Investigating what robots were used for during the coronavirus pandemic provides a de facto work domain analysis that can guide R\&D for future pandemics and possibly future disasters in general. Variations in applications by country may help identify cultural, economic, and policy differences that impact adoption and economic markets.
    \item \textit{When were robots used?} Documenting whether robots were operational before, contemporaneously with, or lagged after the surge for a country captures the overall readiness of robots for a disaster. 
   \item \textit{How did different countries innovate?} Documenting the types of robotics innovations exhibited by individual countries to cope with pandemics can inform research, development, and technology transfer for future disasters. This understanding can be used to craft appropriate policies, including national robotics initiatives. 
   
   \item \textit{Did having a national policy on robotics influence a country's innovation and implementation of robots for COVID-19?} Six countries have national or economic union initiatives in robotics. While those policies differed in scope and size, the existence of a formal commitment to robotics could be expected to foster rapid insertion or adaption for a disaster. Documenting the performance of those countries with initiatives can be useful for creating new or modifying existing policies. 
\end{itemize}

\begin{table}[t]
 \caption{List of countries organized by number of instances of robot use for COVID-19 with the associated
 first date of use. * indicates that the date was inferred.}
 \vspace{0.125in}
 \label{table:countries}
 \label{table:dates}
 \centering
 \includegraphics[width=6.0in]{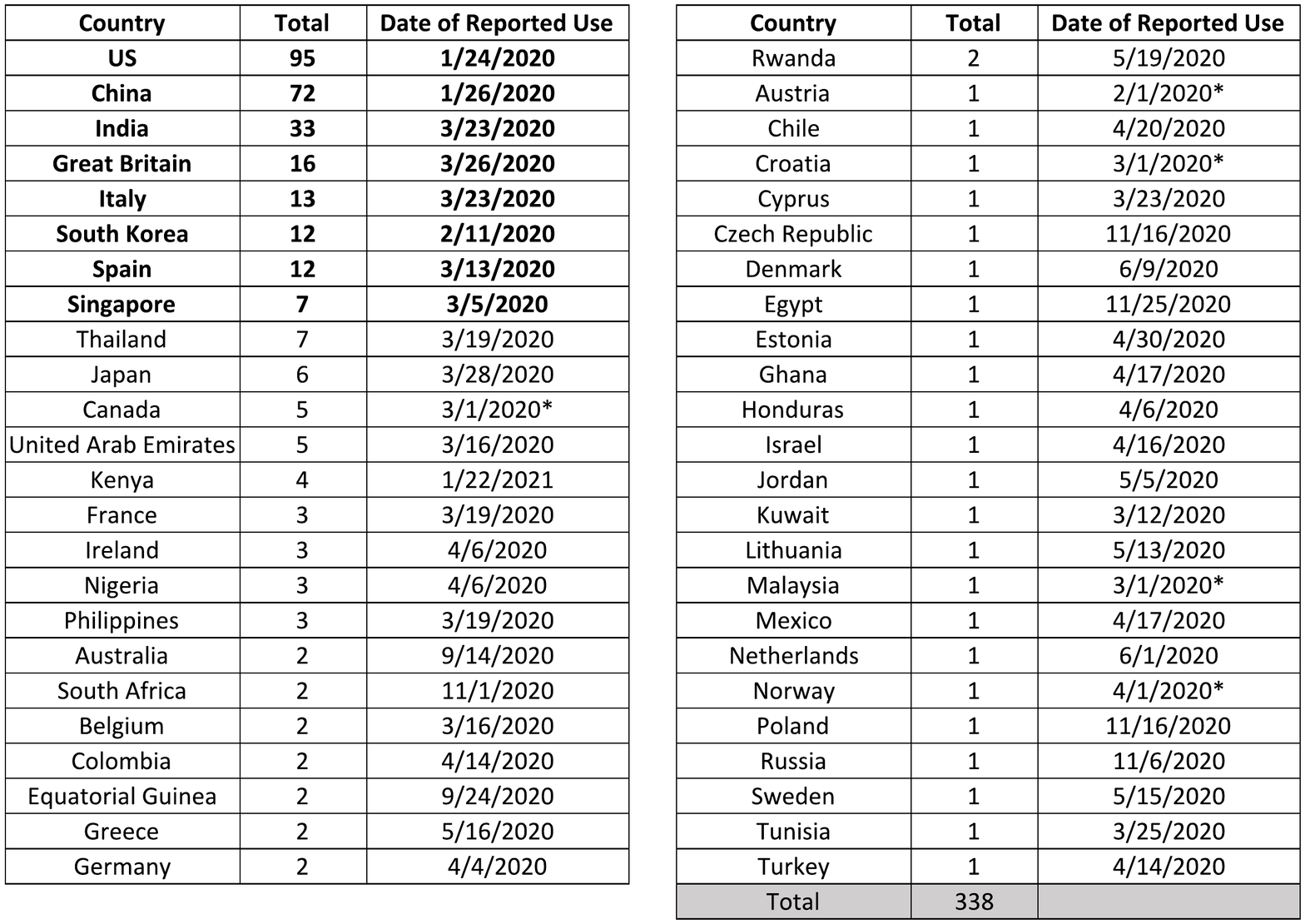}
\end{table}

\begin{figure} [h]
    \centering
    \includegraphics[width=5.0in]{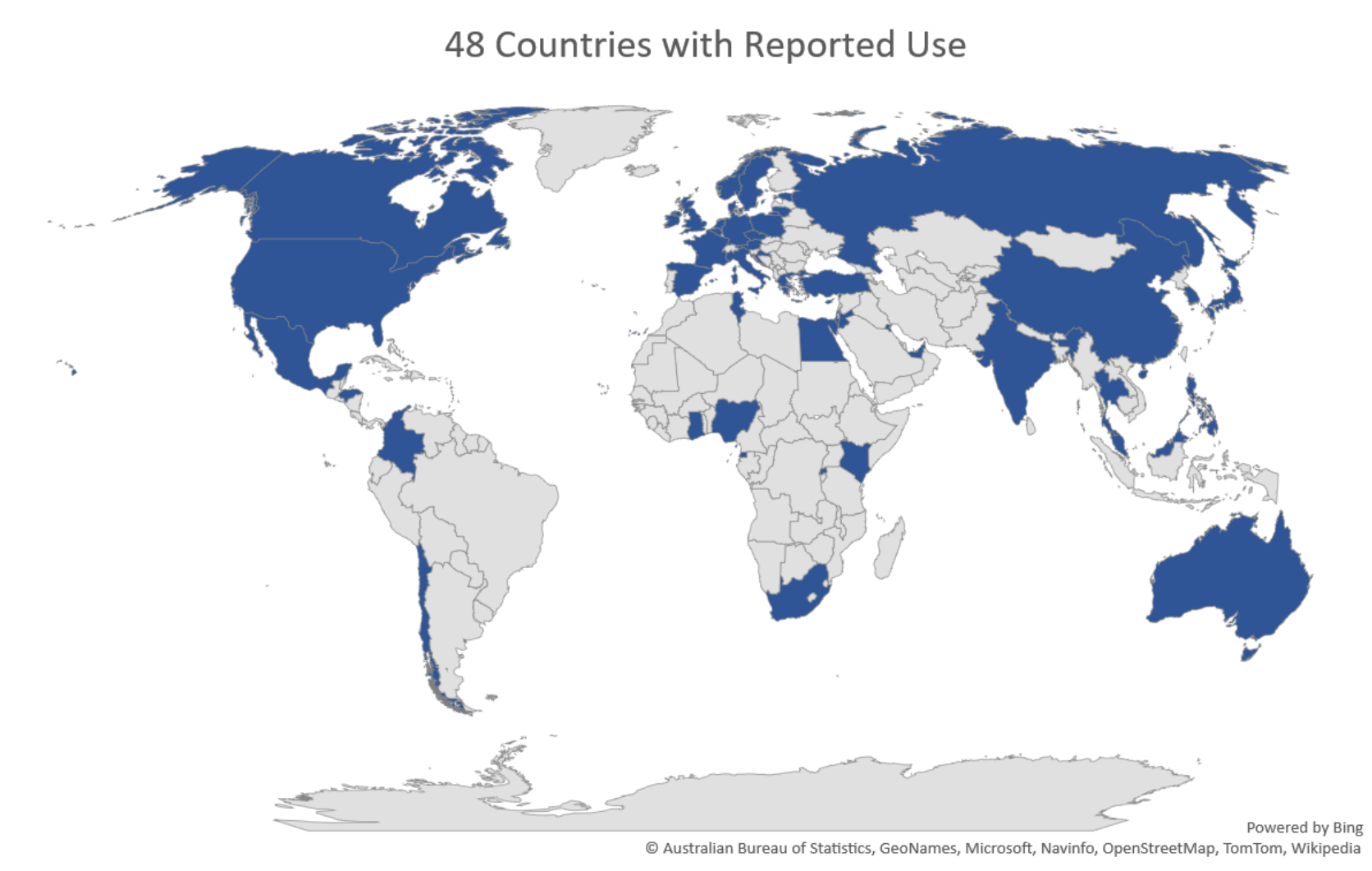}
    \caption{48 countries reporting use of robots for COVID-19 shown in dark blue; map generated through Microsoft Excel.}
    \label{fig:map}
\end{figure}

The analysis in this article is based on the open source Robotics For Infectious Diseases (R4ID) dataset \cite{R4ID}. R4ID captures the use of ground, aerial, and marine robots from the first reported instance, 24 Jan 2020, to 23 Jan 2021, covering a calendar year. 
As of 23 Jan 2021, R4ID contained 338 distinct instances of robots being used explicitly to cope with the COVID-19 pandemic. The instances were from 48 countries in six continents, Africa, Asia, Australia, Europe, North America, and South America. Table~\ref{table:countries} provides a list of countries, while Figure~\ref{fig:map} shows those 48 countries on a map. 

In order to examine international trends in more detail, this article  provides 
summative data for all 48 countries and delves deeper into
the eight countries with the largest number of instances of robot use:
US,
China,
India,
Great Britain,
Italy,
South Korea,
Spain, and
Singapore. Thailand is tied with Singapore for number of instances, 
 but for the purposes of this article and space limitations, Singapore is considered as occupying the eighth position because it has an earlier date for the first use.
Two of these eight countries, Italy and Great Britain are also among the ten countries with the highest per capita incidence of COVID-19 deaths \cite{deaths}. Thus, the top eight provides a tractable number for investigation while representing countries with large COVID-19 impacts and 
sizable deployments of robots. Furthermore, 
six of the eight countries, China, Great Britain, Italy, Spain, South Korea, and the US, have formal robotics initiative programs, possibly providing insights into the impact of national policies. 

The article is organized as follows. It begins with a description of the R4ID dataset and the data collection and analysis methodology. 
Next, each of the following sections explores one of the motivating questions by
presenting relevant data and then extracting findings. 
The article concludes with observations for the robotics and policy communities. 

\section{Data Collection and Analysis Methodology}
\label{sec:methodology}

The robot instances analyzed in this article are extracted from the R4ID database. Each instance documents an operational robot explicitly used for COVID-19, the robot's modality, manufacturer and model, geographic location of use, and date of use. The database is restricted to operational robots, which are defined as those situated
in the work place, not a laboratory or simulation, and in use by end-users, not developers, 
for actual use, not a strictly controlled experiment. This
eliminates many of the speculative robots currently under development cited in
the surveys by \cite{clipper:2020,elavarason:2020,mardani:2020,shen2020,yang20} and allows the analysis in this article to be based on
actual documented use.

The R4ID instances enable the analysis of when robots were used and cumulative frequency plots by country. The extracted instances were clustered into sociotechnical work domains and use cases using the
 constant comparative method of quantitative analysis \cite{glasser:65,glasser:67,merriam:98,ruona:2005}
 to describe what robots were used for internationally. Each robot instance were also rated for technical maturity using the NASA Technical Readiness Assessment process \cite{tra?,tra2} that will be described in more detailed in Section~\ref{sec:tra}. 

The roboticsForInfectiousDiseases database was created by weekly searches of online press reports, social media, and the scientific robotics and medical literature. The searches began in March, 2020, when a pandemic was declared.  English phrases
and keywords (COVID, COVID19, COVID-19 robots, COVID19 Robots, COVID 19 Drone, COVID 19 UAS, COVID Drone, COVID UAS, “COVID-19 and Robots”, “Use of Robots for COVID-19”, “Use of Robots for the present pandemic”, and “COVID-19 Robot uses”) were used. The search often uncovered non-English reports as “robot" and “COVID" are generally expressed as those words regardless of language. In addition, the comments section of the social media posts, and press reports were manually scraped as well to obtain additional links. Every effort was made to determine the date of actual use of a robot, and if that was unknown, it was noted and the date was
considered the day of the report for analysis purposes.
The search discovered the first reported instance of a robot being used for COVID-19 in the world occurring on 24 Jan, 2020, in a hospital in Seattle, Washington, US \cite{VIGNESH2}. The searches were discontinued on 23 Jan, 2021, providing a calendar year of robotics reports. 

The database contains 424 reports leading to 338 distinct instances of use. The 424 reports were filtered to retain only the 400 reports that discussed
actual robots put into operational use explicitly because of COVID-19. The reports often did not
given any indication of the length of operational use or quality of use; therefore, the instances
include situations where the robot was probably a working demonstration rather than placed in regular service. However, this filtering did eliminate reports about robots in laboratories or under development with no explicit date of deployment. The 400 reports contained duplicates, either retweets and 
repostings or a different news agency covering the same robot. These were merged, resulting in 272
entries. Since several news articles described multiple robots or how a model of robot had been used for multiple applications, those entries were split into 338 separate instances of a robot and application tuple.

While every effort has been made to produce a comprehensive, replicable search, there are numerous
limitations of the R4ID dataset. 
For example, the R4ID dataset  is most certainly incomplete and noisy. Robots in service may not have been the subject of press or research reports and thus skipped. 
Also, the use of English keywords may have likely limited discovery, though as noted earlier ``robot" and ``covid" appear in those forms in many non-English postings. 
From both a robotics and epidemiological standpoint, the data is usually coarse. The robotics reports typically did not provide  technical descriptions of the work envelope, though the general characteristics could be inferred from photos and videos,
metrics used to evaluate the performance of the robots, or
the duration of the operation.
However, the volume of reports suggest that it is 
sufficient to detect general trends in robotics and robotics policy and to identify topics for further investigation. 
It is difficult to precisely correlate lockdown and surge dates with robot instances 
as surges and lockdowns were often local rather than national, multiple media sources can be inaccurate, and lockdowns did not occur in some countries. However, examining the trends in cases and deaths for entire countries provides helpful insights.

The 338 instances were iteratively clustered into sociotechnical work domains and use cases using 
the constant comparative method of quantitative analysis \cite{glasser:65,glasser:67}. The grouping into sociotechnical work domains was based on similarities in the stakeholders and regulations impacting implementation, the
end-user skills and expectations, the general work envelope, and the general mission or objective for having robots.

.
Six sociotechnical work domains emerged:
\begin{itemize}
    \item \textsc{Clinical Care}. This refers to hospital and patient related functions that generally occur in structured medical environments with specialized personnel. These activities are regulated by the government and medical insurers who ultimately must approve the use of robots and cost reimbursement. The term \textsc{Clinical Care} is used in its broadest sense and includes intensive care units as well as patient assessment and the diagnosis and treatment processes discussed in robot surveys for medicine such as \cite{dilallo}. 
    \item \textsc{Non-hospital Care.} This covers medical facilities outside of hospital care, such as quarantine camps, nursing homes, and private clinics. These facilities share some of the trained personnel and processes as \textsc{Clinical Care} but operate under different budgets and regulations. 
    \item \textsc{Laboratory and Supply Chain Automation.} This sociotechnical work domain represents the medical support industry which performs specialized functions such as processing tests for coronavirus and supplying hospitals and public health workers. While they are part of the medical response, they are economically and functionally distinct. 
       \item \textsc{Public Safety}. This refers to public health activities carried out by law enforcement or other emergency management agencies. These activities include disinfecting public spaces and enforcing quarantine restrictions by trained personnel. 
    \item \textsc{Continuity of Work and Education}. This sociotechnical work domain encompasses activities by private businesses and educational institutions to maintain operations. While businesses and educational institutions have different funding streams, they share similar work envelopes and have to contend with workers or students who may not be trained to use or interact with robots.
    \item \textsc{Quality of Life.} This sociotechnical work domain refers to the ways in which robots are used to support individuals, either by private companies delivering goods and services or facilitating pandemic-appropriate social interactions. Individuals themselves may use personal robots in new ways to cope with pandemic lockdowns, for example, walking a dog with a small unmanned aerial system. 
\end{itemize}

\begin{figure} [h]
    \centering
    \includegraphics[width=6.5in]{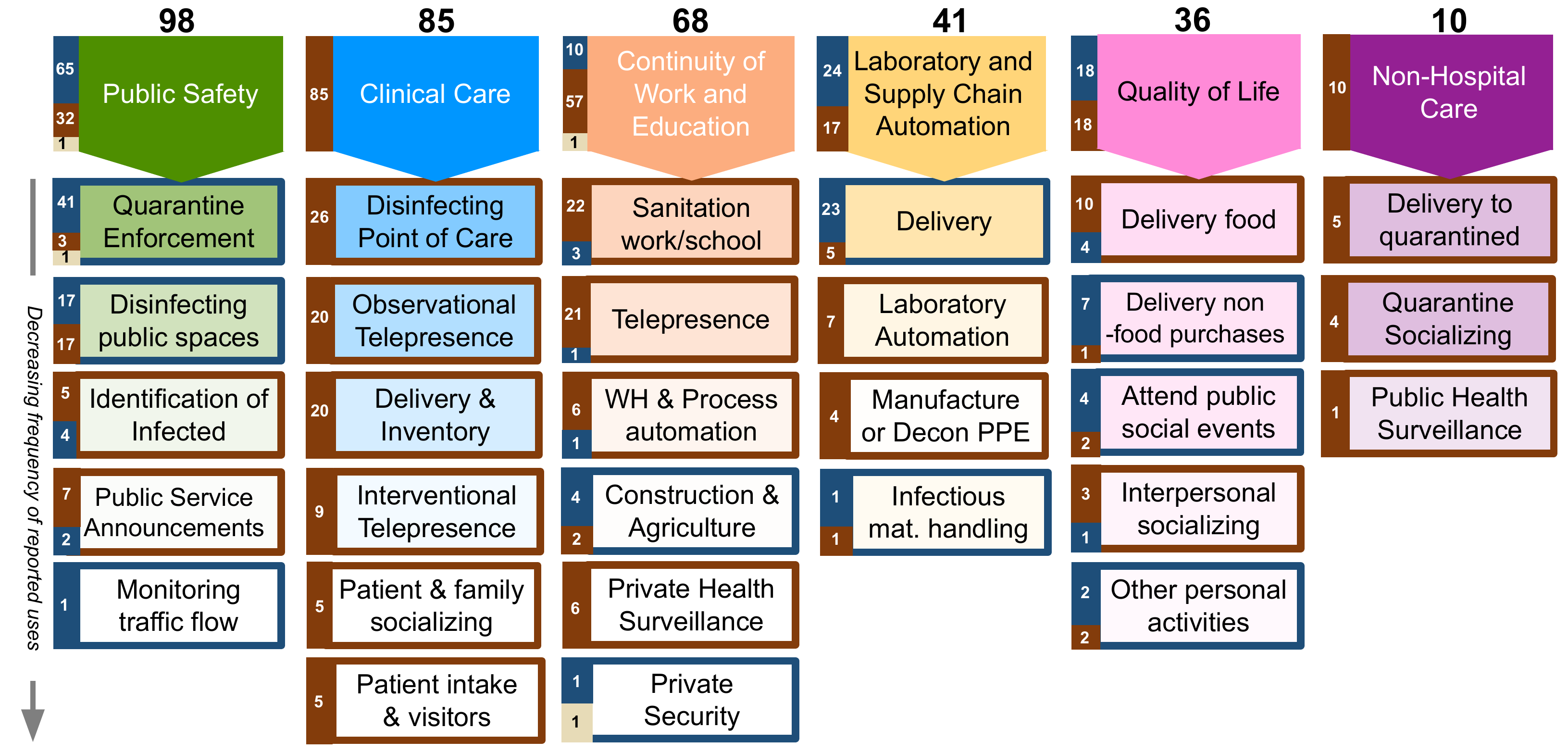}
    \caption{Instances of ground (brown), aerial (blue), and marine (gold) robot by sociotechnical work domain and use case.}
    \label{fig:taxonomy}
\end{figure}

The constant comparative method was also used to cluster the instances within a sociotechnical work domain into use cases. These use cases are shown in Figure~\ref{fig:taxonomy} with largely self-explanatory labels. Perhaps the most confusing labels are for \textsc{Observational Telepresence} and \textsc{Interventional Telepresence} in \textsc{Clinical Care}; observational refers to non-contact interactions such as remote patient assessment while interventional refers to teleoperating a robot that makes physical contact with a patient.  It should be noted that several work domains have a delivery use case. These are considered different use cases for many reasons. Consider robot delivery of medical supplies in a hospital ward (\textsc{Clinical Care}) differs from a small aerial vehicle dropping of a lightweight package of test vials at a laboratory (\textsc{Laboratory and Supply Chain Automation.}), which is in turn different from a robot car dropping off a week's worth of groceries to a surburban household (\textsc{Quality of Life}). Likewise disinfecting outdoor public spaces or large indoor spaces such as covered stadiums (\textsc{Public Safety}) poses different
technical challenges and personnel than disinfecting a hospital room (\textsc{Clinical Care}) or sanitizing a office (\textsc{Continuity of Work and Education}).

Figure~\ref{fig:taxonomy} summarizes the 338 instances by use of robots by sociotechnical work domain, use case, and modality. Note that two earlier versions of this visualization \cite{conversation,murphy:SSRR2020} have appeared previously.
This version has more instances than and due to the content of the instances, the clustering of work domains and use cases changed.
The top row shows the number of instances for that work domain. The work domain icons are ordered by decreasing frequency, left to right. Each icon has 
sidebars decomposing the instances into aerial (117), ground (219), and marine (2) vehicle modalities. Under each domain icon is a column of use cases, with the icon for each use case also containing a sidebar with number of instances by modality.

Note that this criteria for comparison based on sociotechnical work domains and then by use case
produces a different taxonomy from other surveys. 
The taxonomy in this article is most similar to the one by  \cite{cardona:2020}, 
which appears to be based on one of the two earlier iterations of the R4ID database \cite{conversation}. Cardona et al. restricts their survey to operational robots but provides only nine robots, much less than the 338 in this article.
This article's focus on sociotechnical 
work domains is broader that the strictly economic partitioning used by  \cite{elavarason:2020}. The employment of the constant clustering
method resulted in distinctly different clustering of work domains and use cases
from 
the categories in  \cite{shen2020}. That survey surveys 200 robots, some of 
which appear to be laboratory prototypes, and does not provide any motivation for categorization.
Their categorization appears to be by functionality and skewed toward medical applications: diagnosis and screening, disinfection, surgery and telehealth, 
social and care, logistics and manufacturing, and other. In the clustering used by this article,
disinfection is a function that appears as distinct use cases in multiple sociotechnical work domains. For example, disinfecting indoor, sparse hospital rooms which
can be closed off (\textsc{Clinical Care}) poses different robot design considerations
and implementation constraints than robotic disinfection of 
large outdoor public spaces in \textsc{Public Safety}.

\section{What Were Robots Used For?}
\label{sec:uses}


\begin{table}[t]
 \caption{Instances of robots by country for each of the six sociotechnical work domains.}
 \label{table:countries_work}
 \includegraphics[width=6.5in]{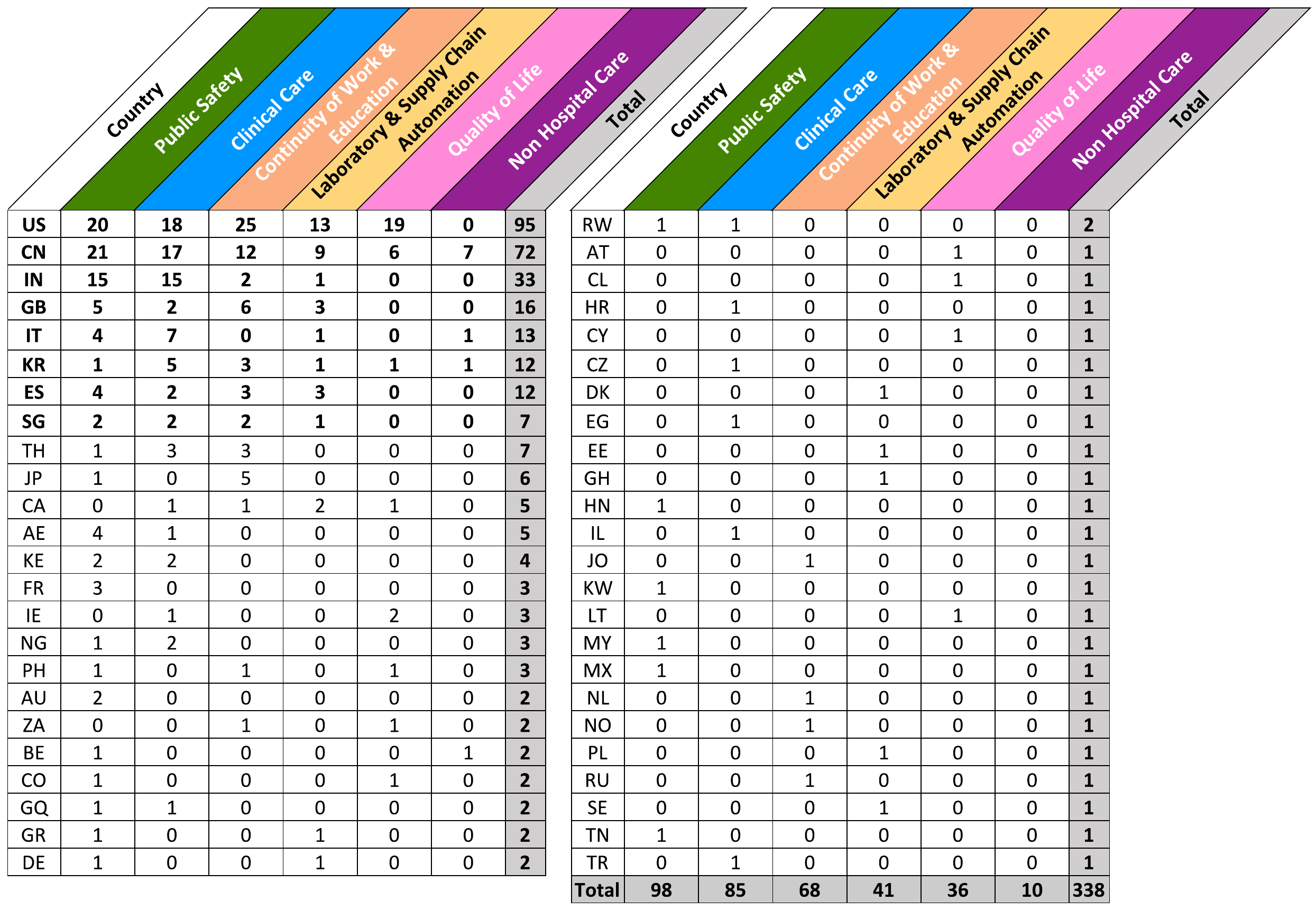}
\end{table}


All three modalities (air, ground, marine) of robots were used for wide variety of applications, spanning six work domains and 29 use cases. While front-line uses of robots for public health were the majority overall and in most countries, there was an almost
equally large use of robots for economic and individual applications, especially in the 
top eight adopters (US, China, India, Great Britain, Italy, South Korea, Spain, and Singapore). 


%
Table \ref{table:countries_work} breaks out the data in Figure~\ref{fig:taxonomy} for each of the 48 countries distributed by work domain. The rows are organized in descending order of total number of instances. 


As would be expected, the majority of instances of use were for aspects of public health, primarily for front-line \textsc{}{Public Safety} (98) and \textsc{Clinical Care} (85), with smaller numbers for \textsc{Laboratory and Supply Chain Automation} (41) and \textsc{Non-Hospital Care} (10). 
However, \textsc{Continuity of Work and Education}, which is driven by businesses and individuals, not by public health or the medical industry, was the third most reported use of robots with 68 instances. 
While each of the 29 use cases differed in some notable way, Figure~\ref{fig:taxonomy} suggests that the majority of the largest use cases could be categorized as robots for disinfection or sanitation to provide effective cleaning without increasing manpower or risking exposure,  telepresence robots to protect users from exposure, and robots delegated for transport and delivery to handle increased surge in demand. 

The majority of robots used were ground vehicles (219), though aerial vehicles (117) were a close second, and two marine vehicles, both unmanned surface vehicles, were used for safety and security applications. Figure~\ref{fig:taxonomy} shows that ground robots were used exclusively for \textsc{Clinical Care} and \textsc{Non-Hospital Care}, while aerial vehicles were used extensively for \textsc{Public Safety}. 

Reported uses were not uniformly distributed by country. As seen in Table \ref{table:countries_work},
only two countries, China and South Korea, had reported instances covering all six work domains. The US had the largest total number of instances but none for Non-Hospital Care, perhaps reflecting  cultural and economic differences in eldercare or a lack of quarantine camps employed in countries such as China. 
17 countries (United Arab Emirates, Kenya, France, Nigeria, Australia, Equatorial Guinea, Rwanda, Croatia, Czech Republic, Egypt, Honduras, Israel, Kuwait, Malaysia, Mexico, Tunisia, Turkey) out of the 48 reported only \textsc{Public Safety} or \textsc{Clinical Care} applications. 
Another 14  countries (South Africa, Austria, Chile, Cyprus, Denmark, Estonia, Ghana, Jordan, Lithuania, Netherlands, Norway, Poland, Russia, Sweden) did not report any use of robots \textsc{Public Safety} or \textsc{Clinical Care}; this could be an artifact of the reporting process or reflect more flexibility in innovation for non-governmental domains. 

These findings suggest:
\begin{itemize}
    \item Research, development, and policies should not limit or conceptualize innovation solely to ``obvious" work domains, in this case to \textsc{Public Safety} or \textsc{Clinical Care}. The volume of instances by businesses and individuals is almost equal to that of public health applications. While the impact of these non-health use cases cannot be assessed from the data, the descriptions suggest that robots were helpful in reducing economic consequences and maintaining society.
\item Even after one year, only 9 instances were reported of interventional telepresence (e.g., requiring physical interaction with a human such as for mouth or nose swabbing) were reported. While such robots are an important topic in fundamental robotics research for medicine, the near term benefits of more mundane and general uses of robots should not be overlooked. 
    \item Unmanned aerial vehicles may be as important as ground robots for a pandemic, but there may be barriers to developing such technologies without a motivating disaster. While reports indicated that emergency waivers of aviation regulations were successfully invoked, the need for waivers suggests that the return of aviation restrictions  may restrict further development and insertion into routine operation.
   
\end{itemize}

\section{When Were Robots Used?}
\label{sec:when}

\begin{figure} [h]
    \centering
    \includegraphics[width=6.25in]{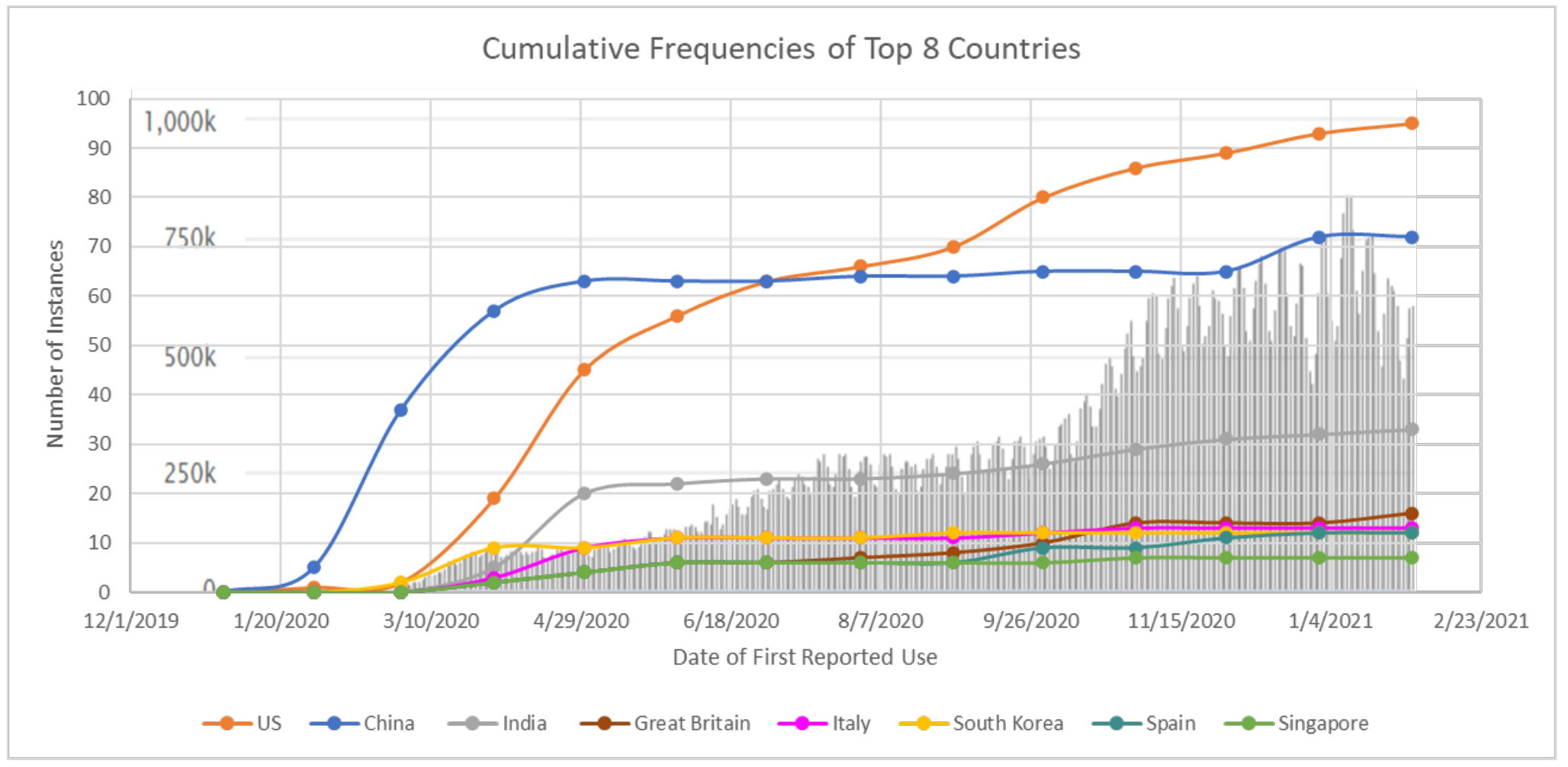}
    \caption{Cumulative number of reports over time from the eight countries superimposed over the worldwide daily case count epidemic curve.}
    \label{fig:cumulative_all}
\end{figure}


Table~\ref{table:dates} captures the first reported date of use of a robot for COVID-19 by country. The World Health Organization declared a pandemic on 11 March, 2020 \cite{who}.
The first eight rows shows countries that had already witnessed 
operational deployment of a robots, with the US on 24 Jan, 2020, was nearly two months before the pandemic was declared. Note that only three of the early adopters were also the countries with the largest deployment, shown in boldface.

Figure~\ref{fig:cumulative_all} illustrates the cumulative number of reports over time from the eight countries superimposed over the worldwide daily case count epidemic curve. 
However, the pandemic reached individual countries at different times
and with dissimilar impacts. Therefore, it is more informative to examine the first use of a robot in a country relative to its
unique epidemic curve. Figure~\ref{fig:cumulative_all} shows the cumulative plot of robot use overlaid with the epidemic curve for the eight countries with the largest reported instances of robot use.  

\begin{figure}
    \centering
     \includegraphics[width=6.5in]{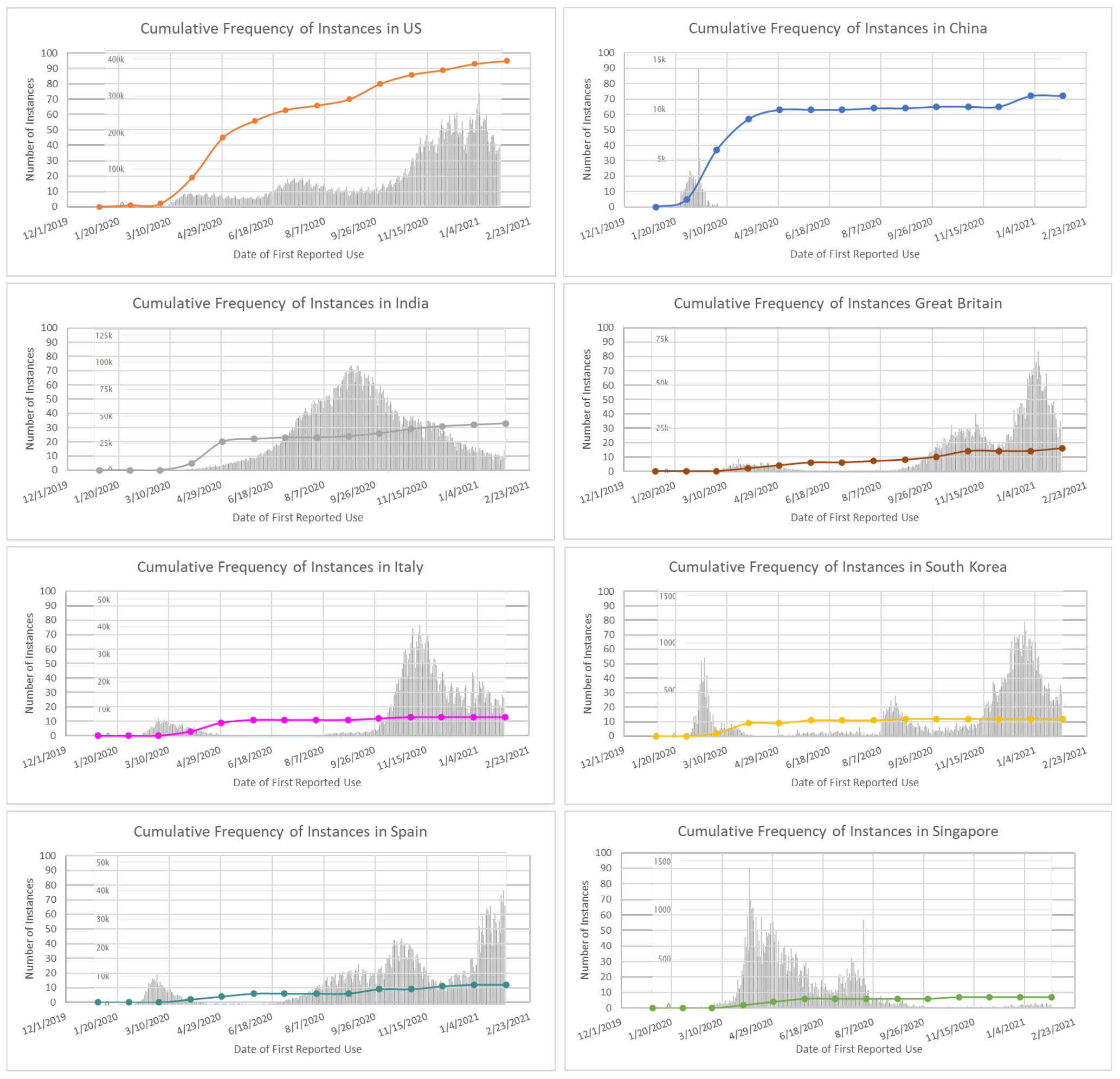}
    \caption{Cumulative plot of robot use overlaid with the epidemic curve for the eight countries with the largest number of instances of robots used for the pandemic.}
    \label{fig:cumulative}
\end{figure}

Figure~\ref{fig:cumulative} and Table~\ref{table:dates} indicate that three of the eight countries,
US, India, and Singapore, 
began deploying robots before their surge, that is, their use led the surge.
Four countries, China, Spain, Great Britain, and South Korea, saw reports of initial robot use
concurrent with their surges. One of the eight exhibited a lag, Italy's first reported use was well after its surge peak. 

Figure~\ref{fig:cumulative} also suggests that early insertions of robots do not guarantee large scale use. For example, the US had the earliest reported operational use of a robot explicitly for COVID-19 but the cumulative plot shows that it was closer to 11 March, 2020, before reports of robots accelerated. Those reports could be misleading as social media could have been focusing on any unfamiliar technology for coping with the rising concern, but if this is a true trend, then it suggests that early successes may not be adequately communicated to other potential  users. 

 Figure~\ref{fig:cumulative_all} also shows that the rate of operationalization varies. All but the US and China showed a rapid rise in reports of robot use followed by a plateau that persisted despite a second (or third) surge. The increasing use rate in the US could be an artifact of the data collection methodology favoring English, however, the data also shows China with a similar curve.

These findings suggest
\begin{itemize}
    \item the application of robotics was mostly reactive, either concurrent with, or lagging, the initial local surge. This may indicate that countries do not have sufficient existing capacity or adopters do not have confidence in robotics except as a last resort; Section~\ref{sec:tra} will discuss this in more depth.
    \item the rate of sustained application of robots varies by country, and could reflect economic and  regulatory frameworks and existing robotics capabilities. 
\end{itemize}

\section{How Did Different Countries Innovate?}
\label{sec:tra}

There are numerous ways to categorize innovation; this article uses the NASA Technical Readiness Assessment (TRA) system and follows NASA's formal process for categorizing innovation by TRA \cite{tra?,tra2}. The TRA system has two advantages. One is that it has a formal decision tree for classifying technology so that the categorization should be uniform and repeatible. The second is that it ranks the robot within 
the human-robot interaction context of its intended use case, not solely by the technical maturity of the robot components. Thus, TRA is well-suited for discussing innovation and inferring why certain robots were adopted. 

 TRA is an expansion of technical readiness levels (TRL) into a broader classification that ranks both the  maturity of a platform (the earlier TRL) and the usability for the work processes \cite{tra?,tra2}. TRA divides readiness into three categories: \textit{Heritage,} \textit{Engineering}, and \textit{New}.   A \textit{Heritage} system is one that is an existing proven technology being applied to a similar mission and work envelope, thus it should not lead to any surprises in usability. An \textit{Engineering} system is a modification of an existing proven technology for a well-defined mission and work envelope. Such a system is highly likely to work and not introduce unintended consequences of increased human cognitive workload or hidden work.  A \textit{New} system either  involves new hardware, software, or a new mission or  notably different work envelope. It is high risk because it is unknown if it will work reliably and without introducing unanticipated demands on the user. 

\begin{table}[h]
\caption{Technical Readiness Assessment of robots by country.}
\label{table:tra_Table}
    \centering
    \includegraphics[width=5.0in]{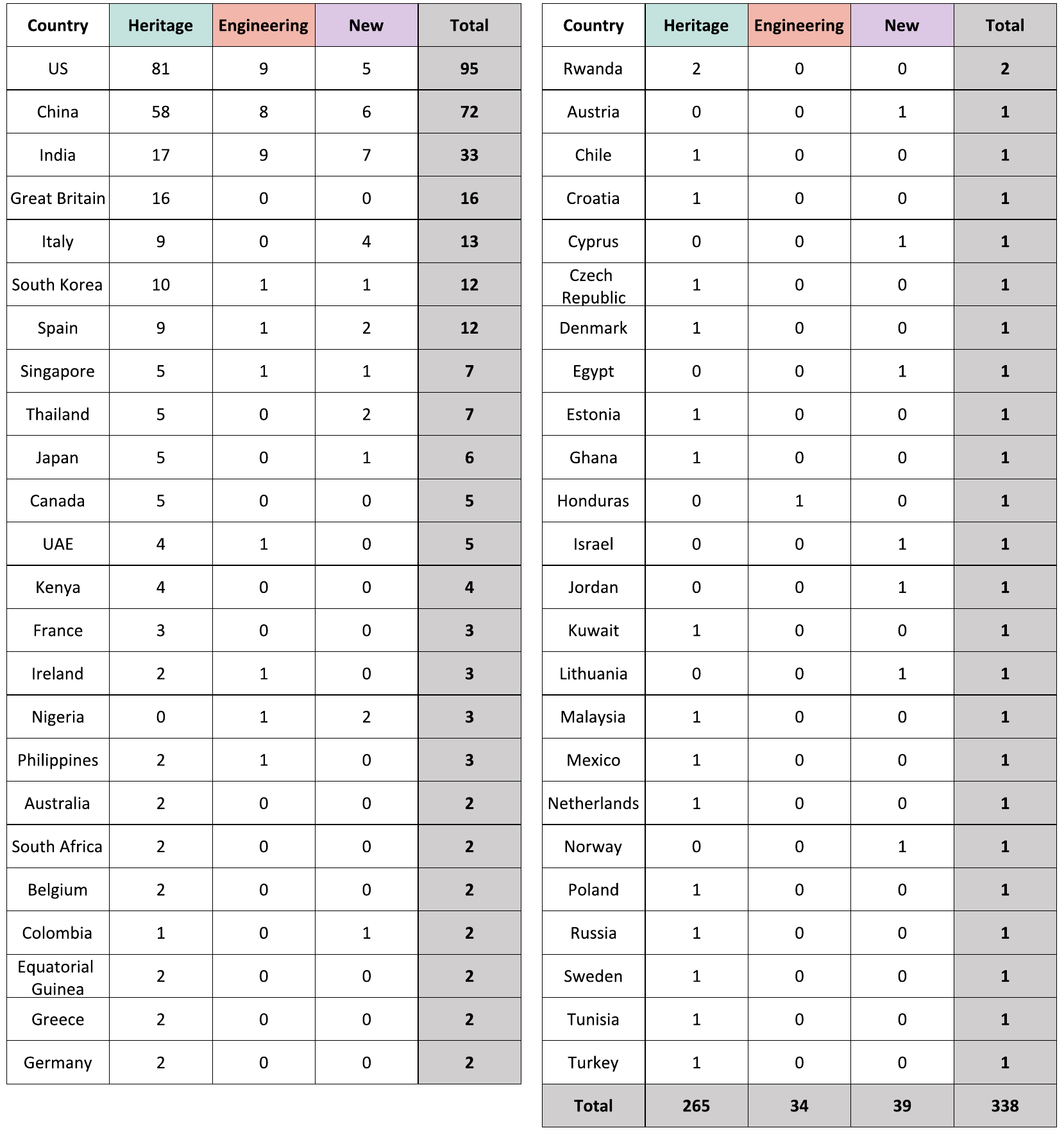}
\end{table}

Following the categorization method in \cite{tra?,tra2}, Table~\ref{table:tra_Table} shows the distribution of Heritage, Engineering, and New robots by country, arranged in descending order of total instances of robots. The overall pattern is for countries to deploy existing robots for established use cases (Heritage) for 78\% of the instances, adapt or repurpose
existing robots for established use cases (Engineering) for 10\%, and create novel robots or address novel
use cases (New) for 12\% of the instances.

\begin{table}[h]
\caption{Comparison of technical readiness of robots by top eight countries.}
\label{table:tra_Table8}
    \centering
    \includegraphics[width=4.5in]{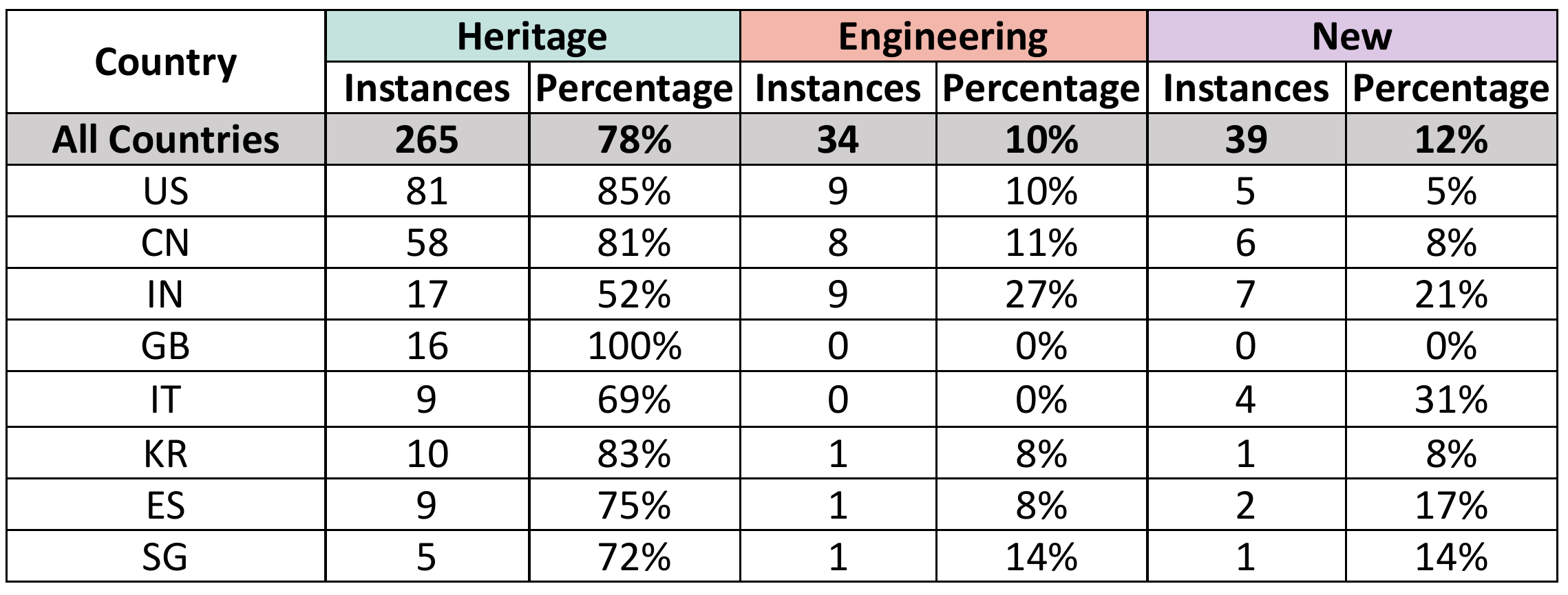}
\end{table}

Table~\ref{table:tra_Table8} details the innovation for the top eight countries. While six of the eight countries generally follow the global trend, India stood out for having a larger percentage of Engineering systems, while Italy had a larger percentage of New systems. An explanation for why India and Italy would differ from the general trend is unclear and worth investigating.

\begin{figure}[h]
    \centering
     \includegraphics[width=6.5in]{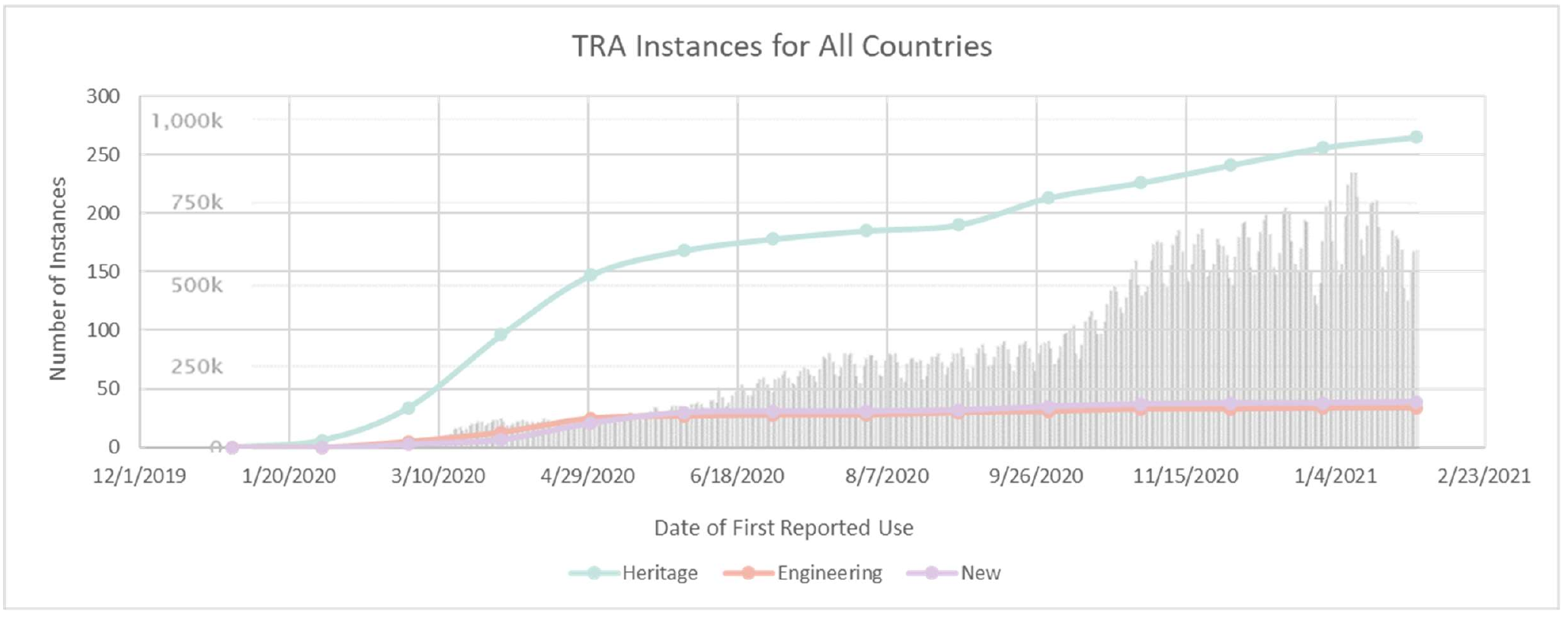}
    \caption{Cumulative plots of robot use by technical readiness for all countries overlaid with the worldwide epidemic curve.}
    \label{fig:tra_time}
\end{figure}

\begin{figure}[h]
    \centering
     \includegraphics[width=6.5in]{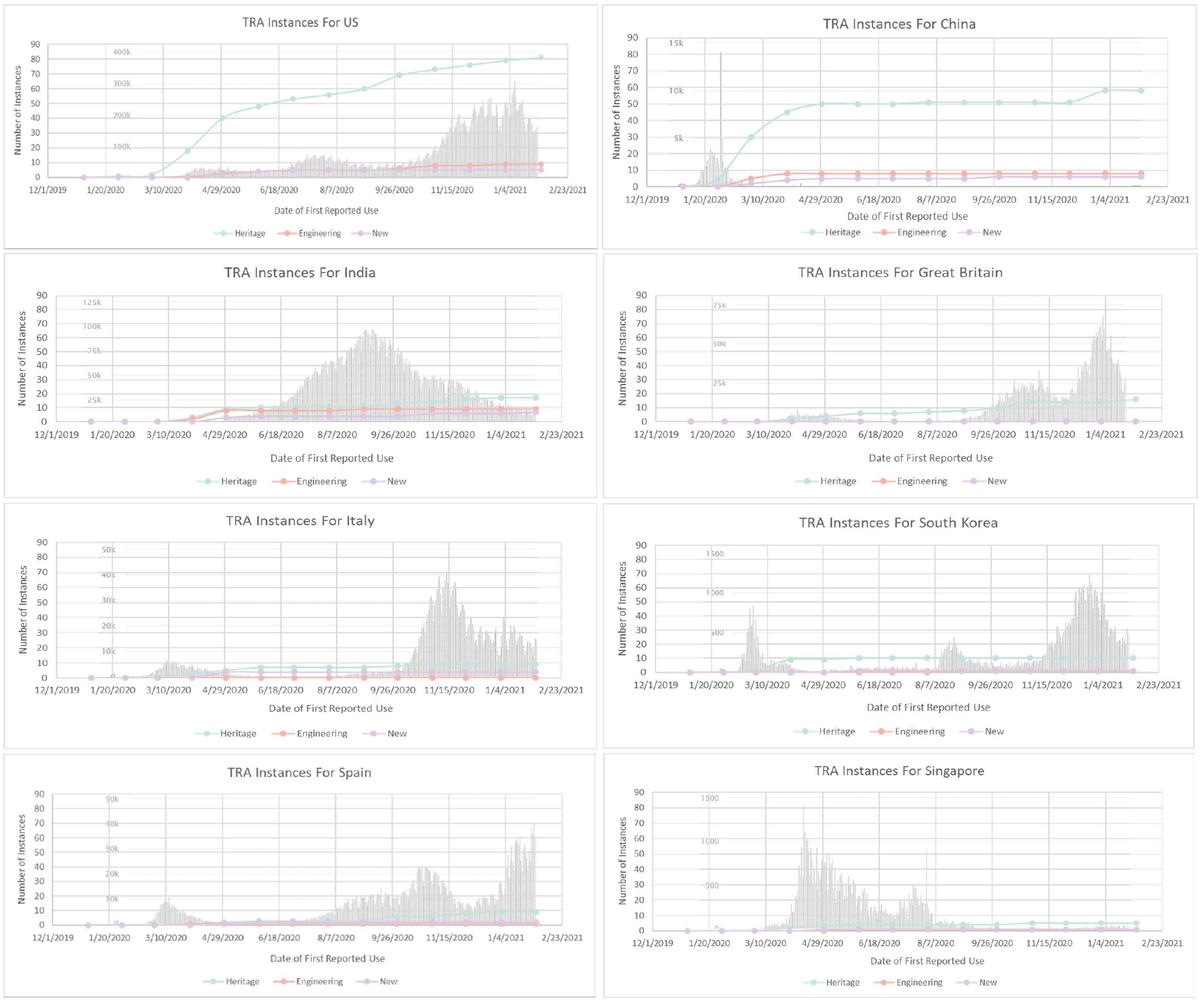}
    \caption{Individual country cumulative plots of robot use by technical readiness overlaid with the epidemic curve for the eight countries with the largest instances of robots for the pandemic.}
    \label{fig:tra_time8}
\end{figure}

The data shows a pattern of innovation over time. Figure~\ref{fig:tra_time} plots the reports of Heritage, Engineering, and New robots by all 48 countries over time. The cumulative graph shows that Heritage robots were deployed first, as would be expected given that users would work with what was available and familiar. However, surprisingly, it indicates that Heritage uses continued to grow over the year and outpace the growth rate of Engineering and New. The data suggests that novel Engineering and New innovations have not been sustained.  Figure~\ref{fig:tra_time8} provides plots of innovation for the top eight adopting countries; the eight, especially the top four, follow the pattern in Figure~\ref{fig:tra_time}. Returning to Table~\ref{table:countries_work} and Table~\ref{table:dates}, it can be seen that smaller countries, such as Rwanda and Poland, typically add Heritage robots; this 
continued addition of Heritage systems may also explain why Heritage reports continue to increase over time. 

\begin{figure}[h]
    \centering
     \includegraphics[width=6.5in]{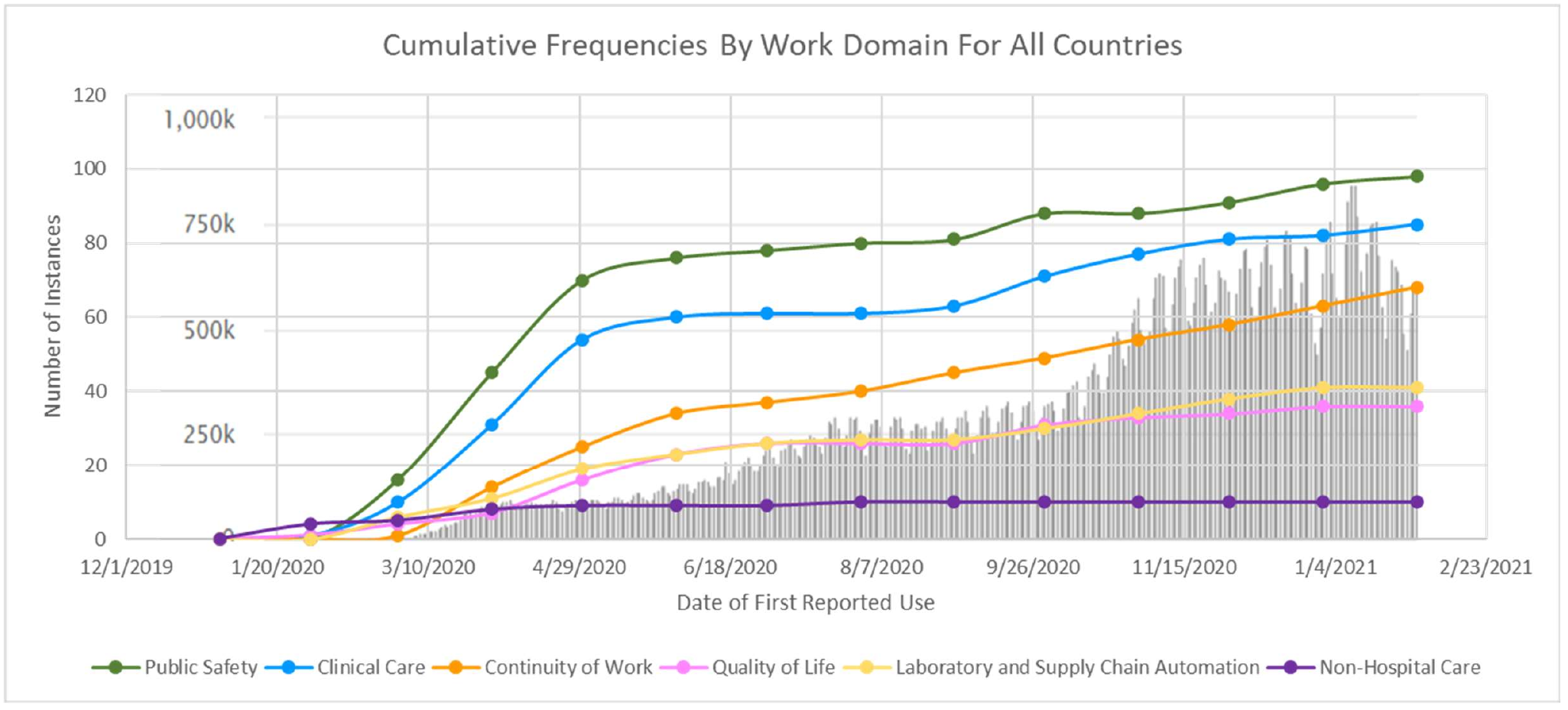}
    \caption{Cumulative plots of robot use by work domain overlaid with the epidemic curve for all countries. }
    \label{fig:work_all}
\end{figure}

\begin{figure}[h]
    \centering
     \includegraphics[width=6.5in]{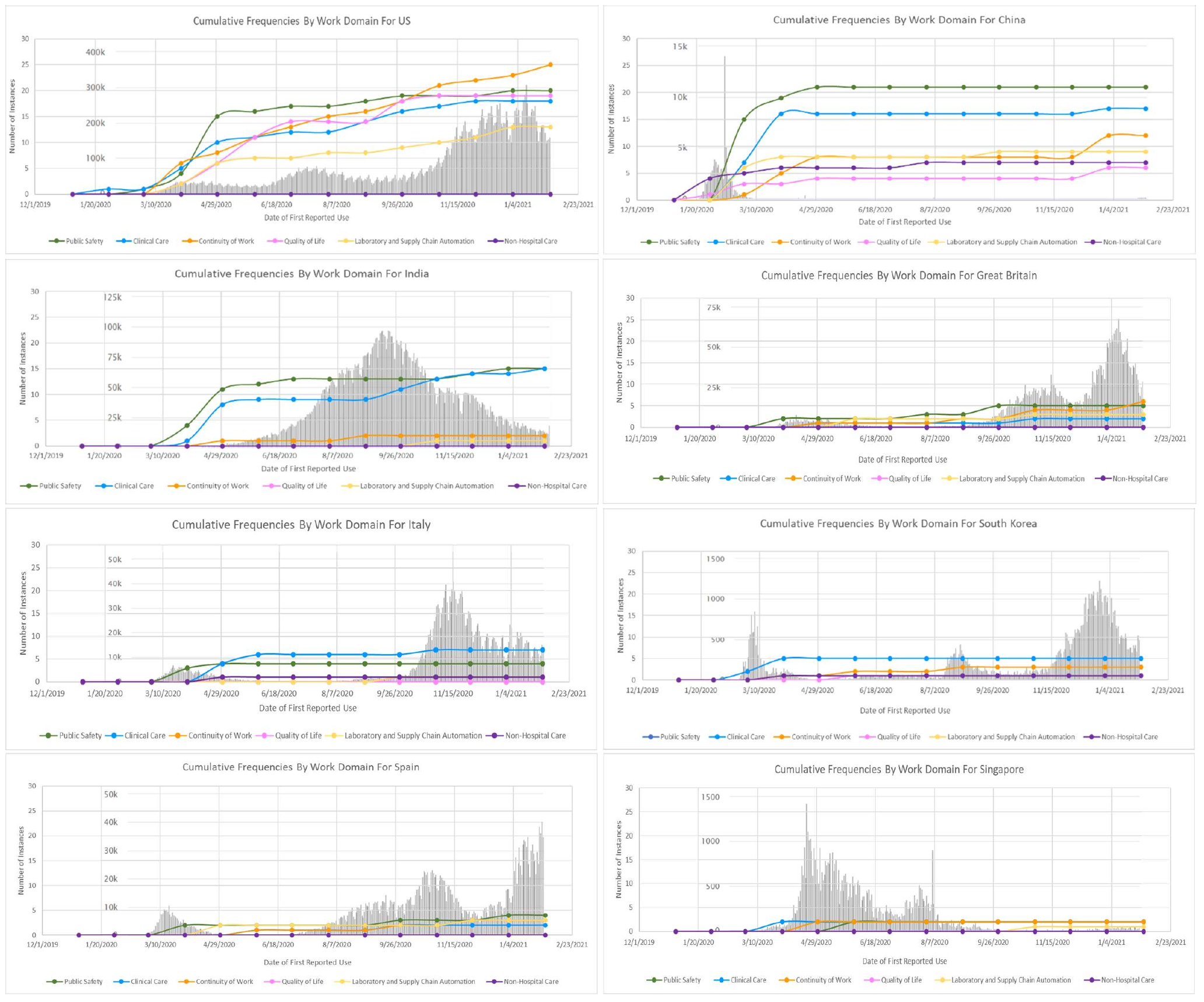}
    \caption{Cumulative plots of robot use by work domain overlaid with the epidemic curve for the eight countries with the largest instances of robots for the pandemic.}
    \label{fig:work_time8}
\end{figure}

The data also shows that robots were reported for multiple sociotechnical work domains from the very beginning of the outbreak and from the beginning of a country's use. Figure~\ref{fig:work_all} illustrates that within approximately a month, robots were being fielded for multiple work domains and not restricted to \textsc{Clinical Care}.  Figure~\ref{fig:work_time8} shows that operational robots in multiple work domains were announced within one month of that country's initial foray.  \textsc{Clinical Care} was not necessarily the first reported use of robots, even though a pandemic is a public health crisis. Three of the top eight countries reported Quality of Life (China) or Public Safety (Spain, Italy) as the first application

The findings on innovation are that:
\begin{itemize}
    \item  Heritage robots, that is, existing robots for known use cases, overwhelming dominated implementations, accounting for 78\% of robots. Engineering and New systems  accounted for only 10\% and 12\% respectively, suggesting that even for a long lasting disaster, users adopt existing systems while developers become immediately engaged in near-term solutions or not at all. 
    \item the global pattern for implementation over time was the immediate use of Heritage robots  and the continued addition of Heritage robots over time. Engineering and New robots experienced an early burst that was not sustained.
    \item innovation did not concentrate on public health priorities; reports of implementation of robots began within one month for all six work domains. This is contrast to the use of robots for high priority public health functions in\textsc{Public Safety} or \textsc{Clinical Care} first, then shifting to  economic concerns captured by \textsc{Continuity of Work and Education} or impacts on individual citizens represented by \textsc{Quality of Life}. This scattershot pattern of applications was the same for the US and China, suggesting that innovation is ultimately independent of government controls of technology development. 

\end{itemize}


\section{Did Having a National Policy on Robotics Influence Innovation?}
\label{sec:initiatives}

Six countries or unions have formal robotics initiatives: 
China, 
the European Union,
Germany (in addition to the European Union), 
Japan,
South Korea, and
the US \cite{initiatives}. 
Of these, Japan and Germany were not in the top eight for reported instances of robotics (recall they are US, China, India, Great Britain, Italy, South Korea, Spain, and Singapore. As part of the EU, Great Britain, Italy, and Spain can be considered as having a robotics initiative during this time period. 
Therefore, of the top eight adopters, six countries had robotics initiatives and two (India, Singapore) did not. 

Having a robotics initiative did appear to impart some advantages. 
As seen in Table~\ref{table:countries_work}, the US, China,  Great Britain, Italy, South Korea, and Spain reported instances for more work domains than other countries, as the average number of domains covered by a country was 2. 
Tables~\ref{table:tra_Table} and \ref{table:tra_Table8} indicates that at least five, the US, China,  Great Britain, South Korea, and Spain,  had a high percentage of Heritage robots to draw upon. This is presumably an outcome of investment in robotics and general societal awareness of robot capabilities. 
However, having a robotics initiative did not ensure that a country visibly deployed robots in a timely manner. Figure~\ref{fig:cumulative} shows that Italy lagged behind its surge. 

The findings suggest:
\begin{itemize}
\item that a national robotics initiative appears helpful in terms of prior availability of existing robots and enabling a breadth of applications. In addition, it may be helpful in creating end-user awareness and acceptance of robots. 
\item The data is less clear as to whether national initiatives impart a clear advantage on rapid innovation, as there were relatively little need for novel robots to meet previously unknown use cases. 
\end{itemize}


\section{Conclusions}

The R4ID dataset show that robots were used for the first year of the COVID-19 outbreak for
an unexpectedly broad set of six work domains and 29 use cases by 48 countries, many of whom are not an advanced economy according the International Monetary Fund \cite{imf}. 
This suggests that
robotics is becoming more mainstream and could auger accelerated adoption by businesses and individuals. 
Creating or expanding existing national robotics initiatives could increase robotics preparedness for future pandemics or disasters. 
It should be emphasized that the R4ID dataset is imperfect and thus any conclusions are speculative and offered for discussion.

Returning to the questions posed in the Introduction, 219 ground, 117 aerial, and 2 marine robots, for a total 338, 
have been documented in use in 48 countries in  Africa, Asia, Australia, Europe, North America, and South America. Robot use was spread
across six distinct work domains: \textsc{Public Safety, Clinical Care, Continuity of Work and Education,
Laboratory and Supply Chain Automation, Quality of Life,} and \textsc{Non-Hospital Care}. 
The majority of robots globally have been used in the \textsc{Public Safety} (98 instances or 29\% of the total) and \textsc{Clinical Care} (85 or 25\%) work domains, which are driven by government and public health policy. 
protect front-line workers (including administrative staff), help cope with surge in demand for medical services
However, the \textsc{Continuity of Work and Education} work domain, at 68 or 20\%, was almost as large as \textsc{Clinical Care}. This suggests that private industry and individuals are acquiring and implementing robots. 
The majority of use cases utilized robots to protect workers, maintain output, or to replace sick workers or enable social distancing. While the majority did not displace workers and would not be 
a threat to jobs long term, there is a possibility that increased robotics for warehouse and production automation will permanently displace workers.

In terms of when robots were used, they were used before, during, and throughout the first year of the pandemic.  Reports show that robots were in service to explicitly cope with the public health crisis in countries such as the US, India, and Singapore before the declaration of the pandemic or their local surges. It appears that the majority of countries, such as China, Spain, Great Britain, and South Korea, deployed robots concurrent with sharp increases in their local surges, while others such as Italy lagged behind their local surge. There is no clear indicator of why some countries were early adopters and others appeared to deploy robots rather late. 


Robotics innovation for the COVID-19 pandemic was primarily a priori innovation. Technically
mature Heritage robots which had been proven in established user cases comprised the largest number of instances, 265 or 78\% of the total. 
Innovation during the pandemic appeared to follow a ``low hanging fruit" strategy, where existing (Heritage) and easy to modify robots (Engineering) are used for established use cases. New innovation
for novel use cases, such as mouth or nose swabbing, or to offer novel robot designs appears to emerge quickly but then slows over time, possibly because of the time it takes to build and program reliable robots with high usability for new applications. It is not possible to know whether New robots would have made a difference, but the high propensity of established use cases suggests that a pandemic does not 
call for new solutions so much as rapidly scaling the availability of existing robots and end-user's
familiarity and trust of such robots. 
 

The existence of a national robotics policy appeared to roughly correlate with a country's use
of robots for COVID-19. Six of the top eight countries in terms of reported instances,
US, China,  Great Britain, Italy, South Korea, and Spain, had national robotics initiatives. 
A national robotics policy appears to be associated with earlier insertion of robots and a greater breadth of work domains that robots are applied to. This positive association with government policy may be because 
many of the instances were for public health work domains which are funded or regulated by the 
government. It could also be that having a national robotics initiative is a reflection
of a country's intrinsic platform availability, talent, and awareness of, or comfort with, robotics. 

It is hoped that this analysis will inform research and development of robots for the next
disaster. The data suggests that national policies are useful and might be 
expanded to incentivize the development of 
 Heritage robots with rapid manufacturing capacity as well as platforms suitable for supporting opportunistic Engineering and New innovations.




\subsubsection*{Acknowledgments}
This material is based upon work supported by the National Science Foundation under Grant IIS-2032729. Any opinions, findings, and conclusions or recommendations expressed in this material are those of the authors and do not necessarily reflect the views of the National Science Foundation.

\bibliographystyle{apalike}
\bibliography{international}

\begin{thebibliography}{}

\bibitem[A man diagnosed with Wuhan coronavirus near Seattle is being treated
  largely by a robot, 2020]{VIGNESH2}
A man diagnosed with Wuhan coronavirus near Seattle is being treated largely by
  a robot (2020).
\newblock
  https://edition.cnn.com/2020/01/23/health/us-wuhan-coronavirus-doctor-interview/index.html.

\bibitem[Cardona et~al., 2020]{cardona:2020}
Cardona, M., Cortez, F., Palacios, A., and Cerros, K. (2020).
\newblock Mobile robots application against covid-19 pandemic.
\newblock In {\em 2020 IEEE ANDESCON}, pages 1--5.

\bibitem[Clipper, 2020]{clipper:2020}
Clipper, B. (2020).
\newblock The influence of the covid-19 pandemic on technology: Adoption in
  health care.
\newblock {\em Nurse Leader}, 18(5):500--503.

\bibitem[Developed Country, IMF Advanced Economies, 2021]{imf}
Developed Country, IMF Advanced Economies (2021).
\newblock https://en.wikipedia.org/wiki/Developed\_country.

\bibitem[DiLallo et~al., 2021]{dilallo}
DiLallo, A., Murphy, R., Krieger, A., Zhu, J., Taylor, R.~H., and Su, H.
  (2021).
\newblock Medical robots for infectious diseases: Lessons and challenges from
  the covid-19 pandemic.
\newblock {\em IEEE Robotics and Automation Magazine}, page to appear.

\bibitem[{Frerking} and {Beauchamp}, 2016]{tra2}
{Frerking}, M.~A. and {Beauchamp}, P.~M. (2016).
\newblock Jpl technology readiness assessment guideline.
\newblock In {\em 2016 IEEE Aerospace Conference}, pages 1--10.

\bibitem[Glaser, 1965]{glasser:65}
Glaser, B.~G. (1965).
\newblock The constant comparative method of qualitative analysis*.
\newblock {\em Social Problems}, 12(4):436--445.

\bibitem[Glaser and Strauss, 1967]{glasser:67}
Glaser, B.~G. and Strauss, A.~L. (1967).
\newblock {\em The Discovery of Grounded Theory: Strategies for Qualitative
  Research}.
\newblock Aldine de Gruyter.

\bibitem[Hirshorn and Jefferies, 2016]{tra?}
Hirshorn, S.~R. and Jefferies, S.~A. (2016).
\newblock Final report of the nasa technology readiness assessment (tra) study
  team).

\bibitem[How nations around the world are investing in robotics research,
  2020]{initiatives}
How nations around the world are investing in robotics research (2020).
\newblock
  https://www.themanufacturer.com/articles/how-nations-around-the-world-are-investing-in-robotics-research/.

\bibitem[Madurai~Elavarasan and Pugazhendhi, 2020]{elavarason:2020}
Madurai~Elavarasan, R. and Pugazhendhi, R. (2020).
\newblock Restructured society and environment: A review on potential
  technological strategies to control the covid-19 pandemic.
\newblock {\em Science of The Total Environment}, 725:138858.

\bibitem[Mardani et~al., 2020]{mardani:2020}
Mardani, A., Saraji, M.~K., Mishra, A.~R., and Rani, P. (2020).
\newblock A novel extended approach under hesitant fuzzy sets to design a
  framework for assessing the key challenges of digital health interventions
  adoption during the covid-19 outbreak.
\newblock {\em Applied Soft Computing}, 96:106613.

\bibitem[Merriam, 1998]{merriam:98}
Merriam, S.~B. (1998).
\newblock {\em Qualitative Research and Case Study Applications in Education. A
  Joint Publication in the Jossey-Bass Education Series and the Jossey-Bass
  Higher Education Series.}
\newblock San Francisco : Jossey-Bass Publishers.

\bibitem[Murphy, 2020]{murphy:SSRR2020}
Murphy, R. (2020).
\newblock How robots are helping with covid-19 and how they can do more in the
  future.
\newblock In {\em 2020 IEEE International Symposium on Safety, Security, and
  Rescue Robotics (SSRR)}, pages 1--1.

\bibitem[Murphy et~al., 2020]{conversation}
Murphy, R., Adams, J., and Gandudi, V. (2020).
\newblock Robots are playing many roles in the coronavirus crisis – and
  offering lessons for future disasters.
\newblock {\em The Conversation}.

\bibitem[Robotics for Infectious Diseases, 2021]{R4ID}
Robotics for Infectious Diseases (2021).
\newblock http://roboticsforinfectiousdiseases.org/index.html.

\bibitem[Ruona, 2005]{ruona:2005}
Ruona, W. E.~A. (2005).
\newblock {\em Analyzing Qualitative Data. Chap. 14 In Research in
  Organizations: Foundations and Methods of Inquiry, edited by Richard A.
  Swanson and Elwood F. Holton, 233-63.}
\newblock San Francisco, CA: Berrett-Koehler.

\bibitem[Shen et~al., 2020]{shen2020}
Shen, Y., Guo, D., Long, F., Mateos, L.~A., Ding, H., Xiu, Z., Hellman, R.~B.,
  King, A., Chen, S., Zhang, C., and Tan, H. (2020).
\newblock Robots under covid-19 pandemic: A comprehensive survey.
\newblock {\em IEEE Access}, pages 1--1.

\bibitem[WHO Director-General's opening remarks at the media briefing on
  COVID-19 - 11 March 2020, 2020]{who}
WHO Director-General's opening remarks at the media briefing on COVID-19 - 11
  March 2020 (2020).
\newblock
  https://www.who.int/director-general/speeches/detail/who-director-general-s-opening-remarks-at-the-media-briefing-on-covid-19---11-march-2020.

\bibitem[Worldometers, 2021]{deaths}
Worldometers (2021).
\newblock https://www.worldometers.info/coronavirus/.

\bibitem[Yang et~al., 2020]{yang20}
Yang, G.-Z., J.~Nelson, B., Murphy, R.~R., Choset, H., Christensen, H.,
  H.~Collins, S., Dario, P., Goldberg, K., Ikuta, K., Jacobstein, N., Kragic,
  D., Taylor, R.~H., and McNutt, M. (2020).
\newblock Combating covid-19—the role of robotics in managing public health
  and infectious diseases.
\newblock {\em Science Robotics}, 5.

\end{thebibliography}

\end{document}